\documentclass{article}

\usepackage{microtype}
\usepackage{graphicx}
\usepackage{subcaption}
\usepackage{booktabs} % for professional tables

\usepackage{hyperref}

\usepackage[preprint]{icml2026}

\usepackage{amsmath}
\usepackage{amssymb}
\usepackage{mathtools}
\usepackage{amsthm}

\definecolor{cvprblue}{rgb}{0.21,0.49,0.74}
\usepackage{makecell}
\usepackage{subcaption}

\usepackage{tikz}
\usepackage{multirow}
\usepackage[table]{xcolor}
\usepackage{graphicx}
\usetikzlibrary{matrix, arrows.meta, positioning}
\usepackage{dsfont}

\usepackage[capitalize,noabbrev]{cleveref}

\theoremstyle{plain}

\theoremstyle{definition}

\theoremstyle{remark}

\usepackage[textsize=tiny]{todonotes}

\icmltitlerunning{CARE: Condition-Aware Representation Regularization for Diffusion Models}

\begin{document}

\twocolumn[
  \icmltitle{CARE: Condition-Aware Representation Regularization for Diffusion Models}



  \icmlsetsymbol{equal}{*}

  \begin{icmlauthorlist}
    \icmlauthor{Fengjia Guo}{equal,yyy}
    \icmlauthor{Zhuoyi Yang}{equal,yyy}
    \icmlauthor{Jie Tang}{yyy}
    % \icmlauthor{Firstname3 Lastname3}{comp}
    % \icmlauthor{}{sch}
  \end{icmlauthorlist}

  \icmlaffiliation{yyy}{Department of Computer Science and Technology, Tsinghua University, Beijing, China}
 
  \icmlcorrespondingauthor{Fengjia Guo}{guofj23@mails.tsinghua.edu.cn}
  \icmlcorrespondingauthor{Jie Tang}{jietang@tsinghua.edu.cn}

  \icmlkeywords{Diffusion Models, Image Generation, Representation Learning}

  \vskip 0.3in
]



\printAffiliationsAndNotice{\icmlEqualContribution}  % no special notice (required even if empty)

\begin{abstract}
Recent advances in diffusion models highlight the importance of representation regularization for improving sample quality and training efficiency. However, commonly used regularization methods often overlook the built-in conditions (such as labels or texts) which directly determine the generation target.  In this work, we demonstrate how conditioning signals affect the feature distribution and introduce the CARE (Condition-Aware REpresentation regularization). CARE is a lightweight plug-and-play regularization framework that dynamically modulates feature distribution based on condition similarity. CARE leverages built-in conditioning signals to judiciously guide the representation space, promoting tighter feature clusters for similar conditions without relying on explicit alignment losses or external supervision. Empirically, CARE consistently improves both visual fidelity and convergence stability across both class-to-image and text-to-image tasks.  On ImageNet, CARE achieves a 19.08\% reduction in FID in 400k training steps, leading to a 3.5$\times$ speed-up. When applied to text-to-image generation, CARE lowers FID by 16.61\% in 200k iterations and improves semantic alignment between generated samples and text prompts. Moreover, CARE can be seamlessly integrated with existing regularization methods, yielding additional performance gains. 
\end{abstract}

\begin{figure}[t]
  \centering
  \includegraphics[width=\linewidth]{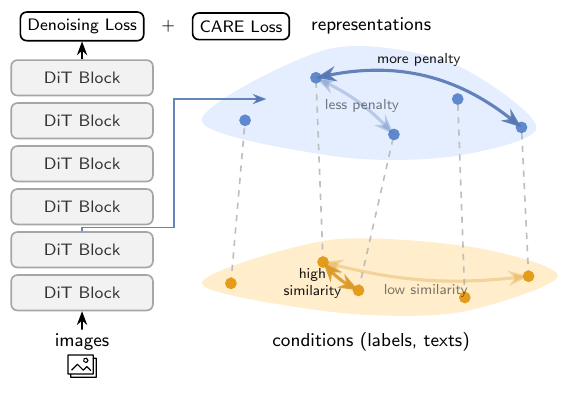}
  \caption{Overview of CARE. Left: DiT pipeline. Right: Representation and condition spaces with CARE regularization. CARE penalizes representation collapse for all sample pairs,
with a similarity-dependent strength based on condition similarity.}
  \label{fig:teaser}
\end{figure}

\begin{figure}[t]
    \centering
    \includegraphics[width=\linewidth]{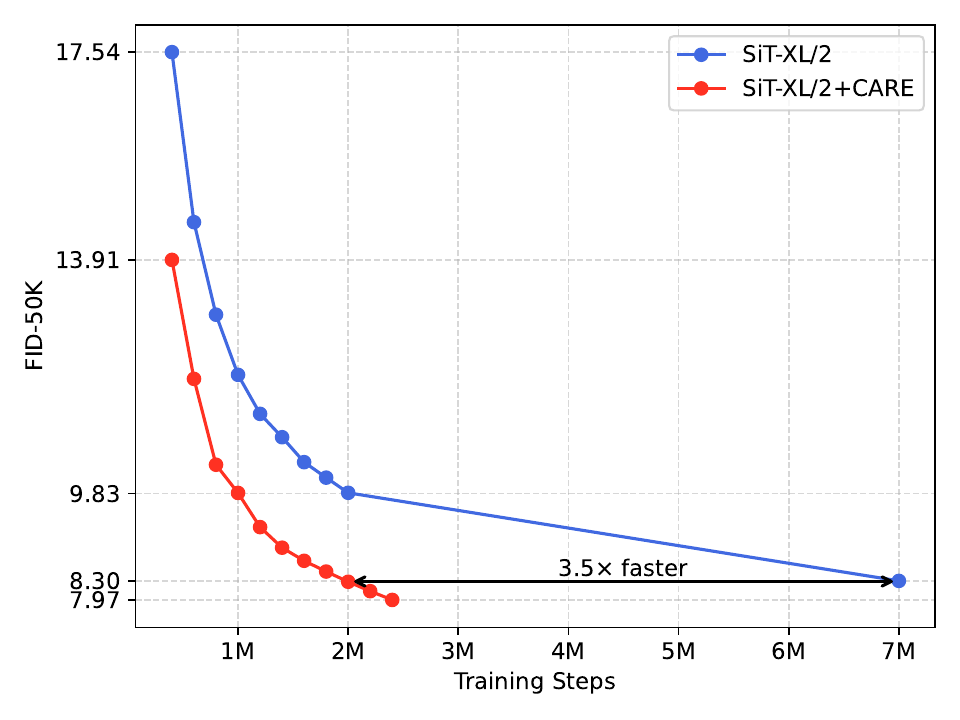}
    \caption{FID-50K on ImageNet 256×256 for SiT-XL/2 with and without CARE. All models are evaluated with SDE sampling, without CFG, using 250 sampling steps. CARE consistently improves FID throughout training.}
    \label{fig:fid}
\end{figure}
\section{Introduction}
Diffusion and flow matching models~\cite{sohl,ddpm,dhariwal2021diffusion, lipman2023flow} have demonstrated remarkable generative capabilities across diverse modalities, including image~\cite{betker2023improving, esser2024scaling}, video\cite{yang2024cogvideox, wan2025wan}, and 3D content\cite{hunyuan3d22025tencent} generation. A widely held view is that representation learning is a key step in generative models training. In particular, recent methods such as REPA~\cite{yu2024repa} and dispersive loss~\cite{wang2025diffusedisperseimagegeneration} have shown that explicitly regularizing internal representations—either by aligning them with vision foundation model\cite{oquab2023dinov2} features or by encouraging feature dispersion from a contrastive perspective—can both accelerate convergence and improve sample quality. However, these approaches do not make use of the semantic information combined with the image/video, such as labels or captions. \textbf{Can those conditions further improve the representation learning in diffusion?} In this work, we give a positive answer via conditional-aware regularization.

\begin{figure}[t]
    \centering
    \includegraphics[width=\linewidth]{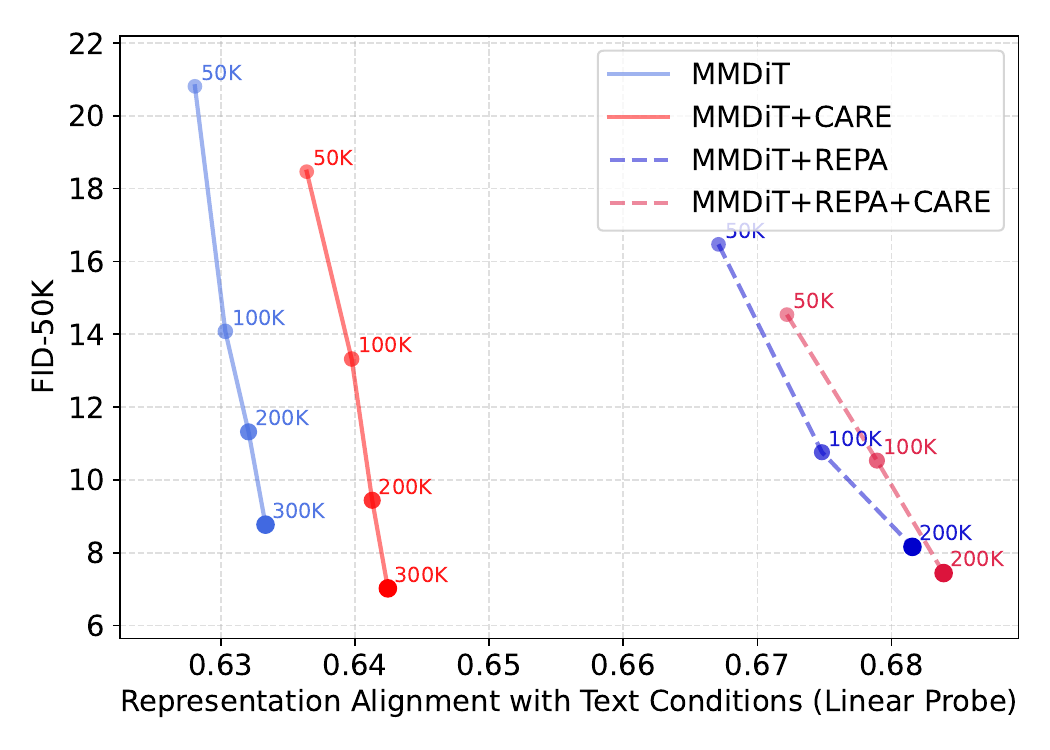}
    \caption{FID-50K versus representation alignment with text conditions, measured by a linear probe on intermediate representations. Across both settings with and without REPA, CARE consistently improves representation--condition alignment while reducing FID.}
    \label{fig:plot-align-t2i}
\end{figure}

To motivate our approach, \cref{fig:plot-align-t2i} illustrates how alignment of intermediate representations with text conditions under linear probe evolves during diffusion training in text-to-image scenarios. 
We observe a strong correlation between alignment and sample quality: as training progresses, representations become more aligned with their conditioning signals, accompanied by a consistent reduction in FID, suggesting that better-structured, condition-aware representation spaces lead to improved generative fidelity.

Yet, existing representation regularizers neglect this conditional structure, leading to a mismatch between the learned feature space and the conditioning signals.
Consequently, samples conditioned on similar inputs may be undesirably scattered, degrading semantic consistency and controllability.

To address this limitation, we introduce \textbf{Conditional-Aware REpresentation Regularization (CARE)}, a simple yet effective plug-and-play loss that injects lightweight supervision into the representation learning objective. Intuitively, CARE leverages semantic conditions to optimize the distribution of representations, requiring no external foundation models, and can be seamlessly combined with existing training objectives or other regularization methods.

In summary, CARE is characterized by three key aspects:

\begin{itemize}
    \item \textbf{An interpretable motivation.}
    CARE is motivated by an information-theoretic interpretation of conditional representation regularization, providing an intuitive explanation of how condition similarity can be incorporated into representation dispersion.
    This perspective is consistent with the Platonic Representation Hypothesis~\cite{huh2024position}, which suggests that representations across modalities tend to exhibit compatible geometric structures.
    \item \textbf{Significant and complementary gains.} CARE consistently improves generation quality across across various conditioned generation tasks. On ImageNet, a class-to-image benchmark, CARE consistently reduces the FID score compared to the baseline steps (see \cref{fig:fid}). On text-to-image benchmarks, CARE yields a further 16.61\% reduction in FID. Notably, CARE also maintains its benefits when combined with representation-alignment techniques. 
    \item \textbf{Enhanced interpretability and structured representations.}
    Beyond performance gains, CARE provides enhanced interpretability of diffusion features. As shown in \cref{fig:plot-align-t2i}, representations regularized by CARE show enhanced alignment with text conditions. 
\end{itemize}

\section{Related Work}
\subsection{Diffusion Models with Regularization}

Recently, diffusion models~\cite{ddpm, song2020score} have become a focal point of research interest due to their outstanding performance in image generation~\cite{ldm, flux2024}.  Recent studies have revealed that explicitly regularizing internal representations of diffusion models can enhance both training efficiency and generation quality. REPA~\cite{yu2024repa} first introduces a feature alignment strategy, encouraging intermediate representations of diffusion models to align with those from powerful pretrained encoders~\cite{oquab2023dinov2}. This alignment facilitates faster convergence and improves the semantic coherence of generated samples. 

Building upon this idea, several extensions have been proposed. For instance, SARA~\cite{chen2025sarastructuraladversarialrepresentation} and REG~\cite{wu2025representationentanglementgenerationtraining} further refine the feature alignment with structural and adversarial views or auxiliary tokens. Meanwhile, dispersive loss~\cite{wang2025diffusedisperseimagegeneration} approaches the problem from a contrastive learning perspective, encouraging feature dispersion to promote uniformity distribution of representations. 

\subsection{Diffusion with Condition-Integrated Objectives}
Conventional approaches integrate conditioning only at the input level, such as by concatenating conditional embeddings with latent variables~\cite{ldm}. In contrast, recent studies seek to explicitly incorporate conditioning signals into the training objective and underlying distributions. 
For example, CAR-Flow~\cite{chen2025carflowconditionawarereparameterizationaligns} introduces a conditional-based reparameterization framework that adjusts both the source and target distributions in flow matching according to conditioning variables. 
Similarly, \citet{issachar2025designingconditionalpriordistribution} proposes to design condition-specific prior distributions for flow-based generative models.  

These methods demonstrate the benefit of incorporating conditional structure directly into the training objective. However, they primarily modify the underlying distributions, and require additional parameters (as in~\cite{chen2025carflowconditionawarereparameterizationaligns}) or multi-stage training (as in~\cite{issachar2025designingconditionalpriordistribution}). 

In contrast, we focus on integrating conditional information into the representation space through a lightweight regularization term, without altering the diffusion process or requiring additional parameterization.  
\section{Method}

\subsection{Overview}
Our goal is to improve the representation quality of conditional diffusion models by explicitly regularizing how features evolve under given conditions. 
While recent works have shown that representation regularization benefits diffusion training, existing approaches are typically condition-agnostic. 
We propose \textbf{Condition-Aware REpresentation Regularization (CARE)}, 
a plug-and-play objective that enforces condition-aware dispersion of intermediate representations 
without relying on external encoders or additional computation.

\subsection{Preliminaries}

Diffusion models~\cite{sohl, ddpm, dhariwal2021diffusion} learn to generate data by reversing a gradual noising process. 
In the continuous-time limit~\cite{song2021denoising}, this process can be formulated as learning a probability flow that transports a simple prior $p_1(\mathbf{x}_1)=\mathcal{N}(0,I)$ toward the data distribution $p_0(\mathbf{x}_0)$.

Flow matching~\cite{lipman2023flow,liu2023rectified-flow} provides a unified framework for this learning process by directly predicting the conditioned velocity field $\mathbf{v}_\theta(\mathbf{x}_t, t, \mathbf{c})$ that satisfies
\begin{equation}\label{eq:fm}
\mathcal{L}_\text{FM} = 
\mathbb{E}_{t,\mathbf{x}_0,\mathbf{x}_1}\!
\left[\left\|\,\mathbf{v}_\theta(\mathbf{x}_t,t,\mathbf{c}) - \frac{\mathrm{d}\mathbf{x}_t}{\mathrm{d}t}\,\right\|_2^2 \right],
\end{equation}
where $\mathbf{x}_t = (1-t)\mathbf{x}_0 + t\mathbf{x}_1$ and $\frac{\mathrm{d}\mathbf{x}_t}{\mathrm{d}t} = \mathbf{x}_1 - \mathbf{x}_0$ denotes the ground-truth velocity under the data–noise coupling.  
The conditioning variable $\mathbf{c}$ (e.g., class label or text embedding) guides the flow toward specific modes of the data distribution $p_0$.

\subsection{CARE: Condition-Aware REpresentation Regularization}

Let $\mathbf{z}_\theta(\mathbf{x}_t, t, \mathbf{c})$ denote the intermediate representation at a given layer of a diffusion model.
Prior work~\cite{yu2024repa} shows that the linear probing accuracy of such representations is positively correlated with sample quality.
We observe a similar phenomenon in text-to-image generation: as training progresses, representations become increasingly aligned with text conditions, accompanied by consistent improvements in generation quality (see \cref{fig:plot-align-t2i}).

This observation resonates with the Platonic Representation Hypothesis~\cite{huh2024position}, which suggests that representations across different modalities tend to organize into compatible geometric structures.
From this perspective, conditioning signals (e.g., class labels or text prompts) and model representations can be viewed as inducing related structures in their respective spaces.
These observations motivate the desideratum that the representation space induced by $\mathbf{z}_\theta$ should preserve conditional structure, such that samples associated with semantically similar conditions exhibit more coherent geometric organization.

Based on this motivation, we introduce \textbf{CARE} (Condition-Aware REpresentation regularization), a lightweight auxiliary objective that explicitly incorporates condition similarity into representation regularization.
CARE does not impose hard constraints on representations; instead, it modulates the strength of repulsive regularization according to condition similarity, encouraging a condition-aware organization of the representation space.
An intuitive interpretation of CARE and its connection to mutual-information-based objectives is provided in \cref{app:interpretation}.

Concretely, we define two functions $s(\cdot,\cdot)$ and $\psi(\cdot)$.
The function $s(\cdot,\cdot)\in[0,1]$ measures the similarity between conditioning signals, while the decreasing function $\psi(\cdot)$ controls the degree to which condition similarity influences representation regularization.
Given a batch of representations and conditions $\{(\mathbf{z}_i,\mathbf{c}_i)\}_{i=1}^N$, CARE is defined as
\begin{equation}\label{eq:care-loss}
\mathcal{L}_{\text{CARE}}
= \mathbb{E}_{i,j}\!\left[
\log \left( \sum_{i,j} \psi\!\left(s(\mathbf{c}_i, \mathbf{c}_j)\right)
\, \phi(\mathbf{z}_i, \mathbf{z}_j) \right)
\right],
\end{equation}
where $\phi(\cdot,\cdot)$ is a kernel function defined in the representation space.

The final training objective augments the original diffusion objective with CARE:
\begin{equation}
\mathcal{L}
=
\mathcal{L}_{\text{FM}} + \lambda\,\mathcal{L}_{\text{CARE}}.
\end{equation}

\subsection{Instantiations}
We set the kernel $\phi$ to be a Gaussian kernel,
\begin{equation}
    \phi(\mathbf{z}_i,\mathbf{z}_j)
    \propto \exp\!\left(- \| \mathbf{z}_i - \mathbf{z}_j \|_2^2 / d \right),
\end{equation}
where $d$ is the dimension of $\mathbf{z}_i,\mathbf{z}_j$.
The division by $d$ provides a simple normalization that stabilizes 
the kernel scale in high-dimensional spaces.

\subsubsection{Class-Conditional Diffusion}\label{sec:instantiation-class-cond}
For class-conditional generation, $\mathbf{c}$ is a discrete label, 
and the similarity function reduces to
$s(\mathbf{c}_i, \mathbf{c}_j) = \mathds{1}[\mathbf{c}_i = \mathbf{c}_j]$.  
We set
$\psi(s) = \alpha \in \left(0,1\right)$ if $s=1$ and $\psi(s)=1$ otherwise. 

\subsubsection{Feature-Conditional Diffusion}\label{sec:instantiation-feature-cond}
For feature-conditional diffusion (e.g., text-to-image generation),  
the condition $\mathbf{c}$ is a continuous embedding, such as a CLIP text feature or a control-image feature.  
We explore two alternative formulations of $s(\cdot, \cdot)$:

\paragraph{(1) Linear cosine scaling.}\label{linear-sim}
We first compute the cosine similarity and linearly rescale it to the range $[0,1]$:
\begin{equation}
s_{\text{linear}}(\mathbf{c}_i, \mathbf{c}_j)
= \frac{1}{2} \frac{\mathbf{c}_i^\top\mathbf{c}_j}
{\|\mathbf{c}_i\|\|\mathbf{c}_j\|} + \frac{1}{2}.
\end{equation}

\paragraph{(2) Softmax-normalized similarity.}\label{softmax-sim}
Alternatively, we normalize similarities across the batch to emphasize relative relationships:
\begin{equation}
s_{\text{softmax}}(\mathbf{c}_i, \mathbf{c}_j)
= \frac{\exp\!\left( \frac{\mathbf{c}_i^\top\mathbf{c}_j}
{\|\mathbf{c}_i\|\|\mathbf{c}_j\|} / \tau_c\right)}
{\sum_k \exp\!\left( \frac{\mathbf{c}_i^\top\mathbf{c}_k}
{\|\mathbf{c}_i\|\|\mathbf{c}_k\|} / \tau_c\right)},
\end{equation}
where the temperature $\tau_c > 0$ controls the sharpness of the similarity distribution. 

For both formulations, we introduce a hyperparameter $\alpha$ in $\psi$ to stay aligned with the formulation in class-conditional setting. 
\begin{equation}\label{eq:psi-t2i}
\psi(s) = s \cdot (\alpha - 1) + 1, 0 < \alpha < 1.  
\end{equation}

Both formulations are compatible with the CARE regularization,  
but lead to different behaviors:  
the linear version is an \emph{absolute} measurement of similarity,  
while the softmax version is a \emph{relative} variant within batch.  
We evaluate both in \cref{ablation-sim-t2i} to analyze how the choice of $s(\cdot, \cdot)$  
affects conditional alignment.
 
\section{Experiments}

\begin{table*}[t]
    \centering
    \caption{CARE consistently improves diffusion models across class-to-image and text-to-image benchmarks.}
    \label{tab:main}
    \begin{tabular}{l|cccccccc}
        \toprule
            \rule{0pt}{13pt} \textbf{Method} & \textbf{Iter.} & \textbf{Sampler} & \textbf{NFEs} & \textbf{CFG scale} & \textbf{FID} $\downarrow$ & \textbf{IS} $\uparrow$ &\textbf{CLIP-T} $\uparrow$\\[5pt]
        \midrule
        \rowcolor{gray!10}
        \multicolumn{8}{c}{\textit{class-to-image generation}}\\
        \midrule
        SiT-B/2 & 400k & SDE & 250 & 1.0 & 33.02 & 43.71 & -  \\
        SiT-B/2 + dispersive loss & 400k & SDE & 250 & 1.0 & 31.37 & 47.85 & - \\
        \rowcolor{blue!8} SiT-B/2 + CARE  & 400k & SDE & 250 & 1.0 & \textbf{29.71} & \textbf{50.10} & - \\
        \midrule
        SiT-B/2 + REPA & 400k & ODE & 250 & 1.0 & 24.31 & 62.23 & - \\
        SiT-B/2 + REPA + dispersive loss & 400k & ODE & 250 & 1.0 & 23.36 & 64.79 & - \\
        \rowcolor{blue!8} SiT-B/2 + REPA + CARE & 400k & ODE & 250 & 1.0 & \textbf{21.64} & \textbf{68.50} & - \\
        \midrule
        SiT-XL/2 & 400k & SDE & 250 & 1.0 & 17.19 &  76.52 & -  \\
        SiT-XL/2 + dispersive loss 
        & 400k & SDE & 250 & 1.0 &  15.57  & 81.68 & - \\
        \rowcolor{blue!8} SiT-XL/2 + CARE  & 400k & SDE & 250 & 1.0 & \textbf{13.91} & \textbf{89.84} & - \\
        \midrule 
        SiT-XL/2~\cite{sit} & 7M & SDE & 250 & 1.0 & 8.26 &  \textbf{131.65} & - \\
        \rowcolor{blue!8} SiT-XL/2 + CARE  & 2.4M & SDE & 250 & 1.0 & \textbf{7.97} & 131.16 & - \\
        \midrule 
        SiT-XL/2 & 400k & SDE & 250 & 1.5 & 5.36 &  167.75 & - \\
        SiT-XL/2 + dispersive loss 
        & 400k & SDE & 250 & 1.5 & 4.74  & 178.81 & - \\
        \rowcolor{blue!8} SiT-XL/2 + CARE & 400k & SDE & 250 & 1.5 & \textbf{4.09} & \textbf{193.76} & - \\
        \midrule
        SiT-XL/2~\cite{sit} & 7M & ODE & 250 & 1.5 & 2.13 &  256.58 & - \\
        \rowcolor{blue!8} SiT-XL/2 + CARE  & 2.4M & ODE & 250 & 1.5 & \textbf{2.10} & \textbf{258.63} & - \\
        \midrule 
        SiT-XL/2 + REPA & 400k & ODE & 250 & 1.0 & 8.81 & 119.61 & - \\
        \rowcolor{blue!8} SiT-XL/2 + REPA + CARE & 400k & ODE & 250 & 1.0 & \textbf{8.31} & \textbf{125.34} & - \\
        \midrule
        \rowcolor{gray!10}
        \multicolumn{8}{c}{\textit{text-to-image generation}}\\
        \midrule
        MMDiT & 200k & ODE & 50 & 2.0 & 11.32 & - &  18.25 \\
        MMDiT + dispersive loss 
        & 200k & ODE & 50 & 2.0 & 9.98  & - & \textbf{18.60} \\
        \rowcolor{blue!8} MMDiT + CARE & 200k & ODE & 50 & 2.0 & \textbf{9.44} & - & 18.54 \\
        \midrule    
        MMDiT + REPA & 200k & ODE & 50 & 2.0 & 8.16 & - & 19.25 \\
        MMDiT + REPA + dispersive loss & 200k & ODE & 50 & 2.0 & 7.80 & - & 19.60 \\
        \rowcolor{blue!8} MMDiT + REPA + CARE & 200k & ODE & 50 & 2.0 & \textbf{7.44} & - & \textbf{19.66} \\
        \midrule 
        MMDiT & 300k & ODE & 50 & 2.0 & 8.77 & - & 18.92 \\
        MMDiT + dispersive loss & 300k & ODE & 50 & 2.0 & 7.44 & - & \textbf{19.21} \\
        \rowcolor{blue!8} MMDiT + CARE & 300k & ODE & 50 & 2.0 & \textbf{7.02} & - & 19.16 \\
        \bottomrule
    \end{tabular}
\end{table*}

\subsection{Experiments Settings}\label{sec:setup}
\noindent\textbf{Class Conditioned Image Generation} 
We conduct experiments on the ImageNet dataset~\cite{imgnet} at 256$\times$256 resolution for class-to-image generation. 
Our experiments are conducted on standard SiT~\cite{sit} models. We strictly follow the original implementations in~\cite{sit} and train the models on the latent space produced by SD-VAE~\cite{ldm}. Sampling is performed using the SDE Euler-Maruyamasampler with 250 steps without classifier-guidance (CFG)~\cite{cfg} by default. For evaluations, we report Fréchet inception distance (FID~\cite{fid}) and inception score (IS~\cite{salimans-inception-score}). 

% We set the batch size to 256 and the loss coefficient $\lambda_{\text{CARE}}$ to 0.25. We apply our method to the last layer of SiT models, which achieves the optimal performance.
We use a batch size of 256 and set the loss coefficient $\lambda_{\text{CARE}}$ to 0.25 by default.
Unless otherwise specified, CARE is applied to the last layer of the SiT model and the condition-agnostic weight $\alpha$ is set as $0.5$.

\noindent\textbf{Text-to-Image Generation}
Our experiments on text-to-image generation are conducted on a 24-layer MMDiT model, consistent with the same setup in~\cite{yu2024repa}. We perform experiments on one of the pretrain datasets of LLaVA~\cite{llava}, which is a subset of the CC3M dataset~\cite{cc3m} containing 595k images at 256$\times$256 resolution and captions relabeled with GPT. We train the MMDiT model following the training settings in~\cite{yu2024repa}. 

The condition inputs of the MMDiT and CARE are generated by of CLIP-L~\cite{clip} text encoder. We apply CARE to the 8th layer of the MMDiT model. For evaluation, we sample 50,000 images from prompts the dataset and report the FID~\cite{fid} and textual CLIP score~\cite{clipscore}. 

\noindent\textbf{Combination with REPA}
For experiments with the combination of REPA~\cite{yu2024repa} in both tasks, we apply our method to the same layer as REPA, and use DINOv2-B~\cite{oquab2023dinov2} as the foundation model for alignment. The final loss becomes
\begin{equation}\label{eq:loss-with-repa}
    \mathcal{L} = \mathcal{L}_{\text{FM}} + \lambda_{\text{REPA}} \mathcal{L}_{\text{REPA}} + \lambda_{\text{CARE}} \mathcal{L}_{\text{CARE}}.     
\end{equation}
The loss weight $\lambda_{\text{REPA}}$ is set to $0.5$ in class-to-image task and 0.25 in text-to-image task. 
Unless otherwise specified, the loss coefficient $\lambda_{\text{CARE}}$ is fixed to 0.25 across all tasks.

\noindent\textbf{Alignment with Text Condition}
To quantify how well intermediate representations align with text conditions,
we adopt a linear probing protocol.
Specifically, we extract model representations by setting $t = 0$, corresponding
to clean inputs, and feed the empty token as the conditional input to remove
explicit conditioning signals during representation extraction.
The resulting representations are split into training and validation sets with
a 9:1 ratio.

We then train a single-layer linear probe to predict text condition embeddings
by minimizing the negative cosine similarity.
The probe is trained with a batch size of 16384 and a learning rate of 0.001.
Following common practice, we report the highest average cosine similarity on
the validation set as the alignment score of the diffusion model.

\subsection{Main Results}

We summarize our empirical findings across class-conditional generation and text-to-image (T2I) tasks. Our proposed \textbf{CARE} (Condition-Aware REpresentation regularization) 
(1) consistently significantly improves generation quality across various benchmarks, 
(2) serves as an effective, complementary regularizer to existing objectives, and 
(3) promotes a stronger conditional semantic structure within the intermediate representation space.

\paragraph{CARE improves generation performance across tasks and models.}
We first evaluate CARE as a standalone regularization objective, denoted as \textsc{CARE}. Comparisons are made against vanilla SiT and MMDiT models, as well as versions regularized by the dispersive loss, a condition-agnostic and parameter-free baseline. Quantitative results are reported in Table~\ref{tab:main}.

\medskip
\noindent\textbf{ImageNet.}
As shown in \cref{fig:fid}, CARE yields consistent performance improvements throughout training. For SiT-B/2 at 400k steps, CARE reduces the FID from 33.02 to \textbf{29.71}. Larger gains are observed on SiT-XL/2: at 400k steps, the FID decreases from 17.19 to \textbf{13.91}, a relative improvement of 19.08\%. When CFG is applied, the improvement becomes substantially larger: the FID further decreases from 5.36 to \textbf{4.09}, corresponding to a 23.69\% relative gain. 
Notably, CARE also significantly improves training efficiency:
our SiT-XL/2 model with CARE trained for 2.4M steps achieves lower FID than the baseline SiT-XL/2 trained for 7M steps, which corresponds to the reported convergence
regime in prior work.

\begin{figure*}[ht]
    \centering
    \includegraphics[width=\linewidth]{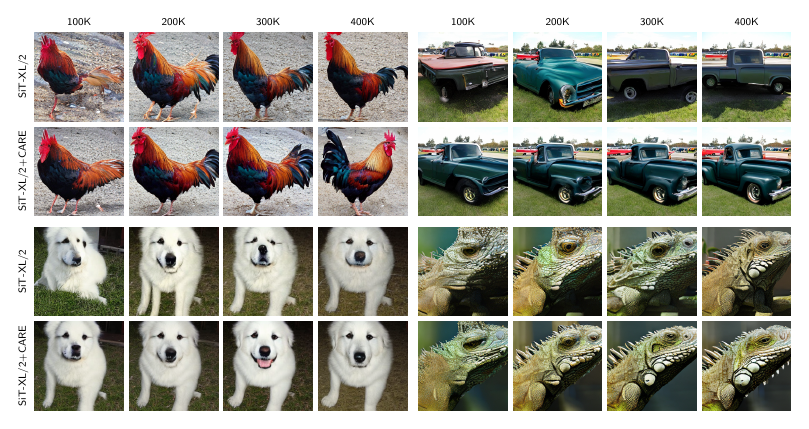}
    \caption{Qualitative comparison on ImageNet 256 $\times$ 256. Both models share the same noise, and use a Euler ODE sampler, 50 steps, and CFG scale $4.0$ for sampling.}
    \label{fig:imgnet-visual}
\end{figure*}

\medskip
\noindent\textbf{Text-to-image tasks.}
CARE also improves both generation quality and alignment with textual conditions. At 200k steps, CARE reduces FID from 11.32 to \textbf{9.44}, a 16.61\% relative improvement. By 300k steps, the FID further decreases from 8.77 to \textbf{7.02}, a 19.95\% improvement. In addition, CARE consistently raises the textual CLIP score throughout training.

\paragraph{CARE produces complementary improvements.}
We combine CARE with REPA to assess whether CARE provides orthogonal benefits to existing representation–regularization approaches. Across both class-to-image and text-to-image settings, the two methods show strongly complementary and synergistic effects.

As shown in \cref{tab:main}, adding CARE on top of REPA leads to further gains across models in class-to-image generation. On SiT-B/2, REPA alone reaches an FID of 24.31, while the combination of REPA and CARE reduces it to \textbf{21.64}, corresponding to a relative improvement of 10.98\%. A similar trend is observed on SiT-XL/2, where pairing CARE with REPA improves the FID from 8.81 to \textbf{8.31}, confirming that CARE provides an additional, orthogonal source of regularization.

\paragraph{CARE promotes better conditional structure of representations.}
To better understand the underlying mechanism, we quantify how well intermediate representations align with conditioning signals in the text-to-image setting, following the evaluation protocol in \cref{sec:setup}. As shown in \cref{fig:plot-align-t2i}, CARE consistently improves condition–representation alignment throughout training, with additional gains when combined with REPA. These results indicate that CARE effectively strengthens the conditional structure of the representation space and provides a clearer semantic organization of learned features.

\subsection{Ablation Studies}

\subsubsection{Effect of Condition-Agnostic Weight $\alpha$}

CARE introduces a key hyperparameter, $\alpha$, which controls the strength of the condition-dependent supervision. When $\alpha = 1.0$, the regularizer becomes fully condition-agnostic. As shown in \cref{tab:cond-weight-imgnet-combined,tab:cond-weight-t2i-combined}, the influence of $\alpha$ exhibits a consistent pattern across both class-to-image without REPA and text-to-image experiments. Moderate values such as $\alpha = 0.5$ generally yield the best trade-off, while the purely condition-agnostic variant tends to perform the worst.

When CARE is combined with REPA on class-conditioned ImageNet, the optimal choice of $\alpha$ shifts toward smaller values. In particular, $\alpha$ in the range of $0.01$ to $0.001$ leads to the strongest improvements. We posit that the presence of external supervision in REPA amplifies the benefit of smaller $\alpha$, as weaker condition-agnosticity encourages the representation to more directly align with REPA's optimization objective.

\begin{table}[h]
    \centering
    \caption{Effect of $\alpha$ on ImageNet with and without REPA. Evaluated using an ODE sampler for 250 steps (without CFG).}
    \label{tab:cond-weight-imgnet-combined}
    \small
    \begin{tabular}{@{}c|cc|cc@{}}
        \toprule
        % \multirow{2}{*}{$\mathbf{\alpha}$} 
        & \multicolumn{2}{c|}{\textbf{w/o REPA}} 
        & \multicolumn{2}{c}{\textbf{w/ REPA}} \\
        $\boldsymbol{\alpha}$ 
        &\textbf{ FID} $\downarrow$ & \textbf{IS} $\uparrow$ 
        & \textbf{FID} $\downarrow$ & \textbf{IS} $\uparrow$ \\
        \midrule
        baseline 
        & 34.84 & 41.53 
        & 24.31 & 62.23 \\
        1.0 
        & 32.79 & 44.80 
        & 23.36 & 64.79 \\
        0.5 
        & \textbf{30.91} & \textbf{47.22} 
        & 23.28 & 64.88 \\
        0.25 
        & 31.55 & 46.17 
        & 23.18 & 64.31 \\
        0.05 
        & -- & -- 
        & 22.25 & 67.33 \\
        0.01 
        & -- & -- 
        & \textbf{21.64} & \underline{68.50} \\
        0.001 
        & -- & -- 
        & \underline{21.83} & \textbf{69.35} \\
        \bottomrule
    \end{tabular}
\end{table}

\begin{table}[h]
    \centering
    \caption{Effect of the $\alpha$ parameter on text-to-image generation, with and without REPA. 
    All results are evaluated using an ODE sampler (50 steps) with CFG scale $=2.0$.}
    \label{tab:cond-weight-t2i-combined}
    \begin{tabular}{@{}c|cc|cc@{}}
        \toprule
        & \multicolumn{2}{c|}{\textbf{w/o REPA}} 
        & \multicolumn{2}{c}{\textbf{w/ REPA}} \\
        $ \boldsymbol\alpha$ 
        & FID $\downarrow$ & CLIP-T $\uparrow$ 
        & FID $\downarrow$ & CLIP-T $\uparrow$ \\
        \midrule
        baseline 
        & 11.32 & 18.25 
        & 8.16 & 19.25 \\
        1.0 
        & 9.98 & \underline{18.60} 
        & 7.80 & 19.60 \\
        0.75 
        & 9.77 & \textbf{18.62} 
        & 8.12 & 19.62 \\
        0.5 
        & \textbf{9.44} & 18.54 
        & \textbf{7.44} & \underline{19.66} \\
        0.25 
        & \underline{9.69} & 18.53 
        & \underline{7.67} & \textbf{19.69} \\
        \bottomrule
    \end{tabular}
\end{table}

\subsubsection{Effect of Similarity Measurement}\label{ablation-sim-t2i}

We investigate two formulations of the similarity function described in \cref{sec:instantiation-feature-cond}: 
\textbf{linear} and \textbf{softmax}. 
These two variants represent \emph{absolute} and \emph{relative} similarity measurements, respectively. 

As shown in \cref{tab:ablation-similarity-t2i}, both formulations yield significant gains in sample quality (FID) and condition alignment (CLIP-T) compared to the baseline.
The linear measure, which directly evaluates feature affinity, provides slightly better overall sample quality, 
while the softmax form, emphasizing relative contextual similarity, enhances condition alignment.
These results suggest that CARE’s effectiveness is robust across different similarity metrics.

\begin{table}[h]
    \caption{Ablation results of similarity measure in text-to-image generation. Evaluated using an ODE sampler for 50 steps \textit{with} CFG scale $=2.0$}
    \label{tab:ablation-similarity-t2i}
    \centering
    \begin{tabular}{@{}ccccc@{}}
        \toprule
        $ \boldsymbol{\alpha} $ & $\boldsymbol{s(\cdot)}$ 
        & \textbf{FID} $\downarrow$
        & \textbf{CLIP-T} $\uparrow$
        \\
        \midrule

        \multicolumn{2}{c}{baseline \textit{w/o CARE}} & 11.32 & 18.25 \\
        \midrule

        \multirow{3}{*}{0.25}
        & linear               & 9.69 & 18.53 \\
        & softmax $(\tau_c = 1)$   & 9.76 & \underline{18.62} \\
        & softmax $(\tau_c = 0.5)$ & 9.77 & 18.55 \\
        \midrule

        \multirow{3}{*}{0.5}
        & linear               & \textbf{9.44} & 18.54 \\
        & softmax $(\tau_c = 1)$   & 9.88 & 18.57 \\
        & softmax $(\tau_c = 0.5)$ & \underline{9.58} & \textbf{18.64} \\
        \bottomrule
    \end{tabular}
\end{table}

\subsubsection{Effect of Loss Coefficient $\lambda_{\text{CARE}}$}

We investigate the impact of the loss coefficient $\lambda_{\text{CARE}}$ on ImageNet $256\times256$, both with and without REPA. As shown in \cref{tab:abl-lambda-care}, introducing CARE with a moderate coefficient consistently improves generation quality over the baseline. Without REPA, setting $\lambda_{\text{CARE}}=0.25$ leads to a substantial improvement, reducing FID from 34.84 to 30.91 while increasing IS from 41.53 to 47.22. Increasing the coefficient to $0.5$ results in slightly degraded performance, indicating that excessively strong regularization may begin to interfere with the original diffusion training objective.

When combined with REPA, both $\lambda_{\text{CARE}}=0.25$ and $0.5$ further improve upon the REPA baseline. In particular, $\lambda_{\text{CARE}}=0.5$ achieves the best FID (21.57), while $\lambda_{\text{CARE}}=0.25$ yields the highest IS (68.50). Overall, these results suggest that CARE is not overly sensitive to the precise choice of $\lambda_{\text{CARE}}$, and relatively small coefficients are sufficient to obtain significant gains. 

\begin{table}[h]
    \centering
    \caption{Effect of the loss coefficient $\lambda_{\text{CARE}}$ on ImageNet 256 $\times$ 256, with and without REPA. All results are evaluated using an ODE samplers for 250 steps (without CFG). }
    \label{tab:abl-lambda-care}
    \begin{tabular}{@{}c|cc|cc@{}}
        \toprule
        & \multicolumn{2}{c|}{\textbf{w/o REPA}} 
        & \multicolumn{2}{c}{\textbf{w/ REPA}} \\
        $ \boldsymbol\lambda_{\textbf{CARE}}$ 
        & FID $\downarrow$ & IS $\uparrow$ 
        & FID $\downarrow$ & IS $\uparrow$ \\
        \midrule
        baseline 
        & 34.84 & 41.53 
        & 24.31 & 62.23 \\
        0.25 
        & \textbf{30.91} & \textbf{47.22} 
        & \underline{21.64} & \textbf{68.50} \\
        0.5 
        & \underline{32.27} & \underline{45.25}
        & \textbf{21.57} & \underline{68.35} \\
        \bottomrule
    \end{tabular}
\end{table}

\subsubsection{Effect of Injection Depth}

We study the effect of injecting CARE at different intermediate layers of the diffusion model.
As shown in \cref{tab:abl-injection-layer}, injecting CARE at deeper layers consistently leads to better generation quality, reflected by both lower FID and higher IS.

Specifically, applying CARE at shallow layers yields only marginal improvements over the baseline. Injecting CARE at layer 12, corresponding to the final transformer block in SiT-B/2, achieves the best results. 

We attribute this trend to the fact that CARE acts directly on the representations used for conditional generation.
When injected at deeper layers, CARE more directly influences the representations that are propagated to the output, making its regularization effect more effective.
In contrast, regularization applied at earlier layers may be partially attenuated by subsequent transformations.

\begin{table}[h]
    \centering
    \caption{Effect of CARE injection layer on ImageNet 256 $\times$ 256  using SiT-B/2 (a 12-layer model). All results are evaluated using an ODE sampler for 250 steps (without CFG).}
    \label{tab:abl-injection-layer}
    \begin{tabular}{@{}c|cc@{}}
        \toprule
        Layer & FID $\downarrow$ & IS $\uparrow$ \\
        \midrule
         baseline & 34.84 & 41.53 \\
         4 & 34.46 & 42.61 \\
         8 & \underline{32.69} & \underline{44.88} \\
         12 & \textbf{30.91} & \textbf{47.22} \\
        \bottomrule
    \end{tabular}
\end{table}

\subsubsection{Enforcing Distinct Labels within Local Batch}\label{sec:ablation-batch-sampler}

To examine whether distancing intra-class representations contributes to effective representation regularization, we conduct an ablation study where each local batch used for the regularization loss contains samples from distinct class labels on ImageNet $256\times256$.
From a contrastive learning perspective, this ensures that no two samples in the batch share the same label, functioning as false negative cancellation~\cite{Huynh_2022_WACV_false_negative_cancellation}. 

\cref{tab:ablation-batch-sampler} shows that this modification yields a moderate improvement for the standard dispersive loss, reducing FID from 32.79 to 32.07.This indicates that avoiding false negatives indeed benefits representation learning.However, the improvement remains smaller than that of CARE,
which further achieves an FID of 30.94.

We attribute this to the design of CARE, which does not simply avoid false negatives
but instead learns the relative structure of representations across conditions - adding a relatively small penalty for clustering of representations with similar conditions. 

\begin{table}[h]
  \caption{Ablation results on enforcing label distinctness within local batch (\textbf{d-sampler}). Evaluated using an ODE sampler for 250 steps \textit{without} CFG}
  \label{tab:ablation-batch-sampler}
  \centering
  \begin{tabular}{@{}lccc@{}}
    \toprule
    \textbf{Method} & \textbf{Iter.} & \textbf{FID} $\downarrow$ & \textbf{IS} $\uparrow$ \\
    \midrule
    SiT-B/2 + disp. loss & 400k & 32.79 & 44.80 \\
    SiT-B/2 + disp. loss + d-sampler & 400k & \underline{32.07} & \underline{45.50} \\
    SiT-B/2 + CARE & 400k & \textbf{30.94} & \textbf{47.22} \\
    \bottomrule
  \end{tabular}
\end{table}

\subsection{Generalizing to other supervision signals}
The framework of CARE can be generalized to scenarios where other supervision signals are available. For instance, we replaced the text conditions in CARE with the image class tokens of the DINOv2 model. When conducting experiments on text-image generation combined with REPA, this modification further improves the FID from \textbf{7.44} to \textbf{7.27} compare to CARE with text class token, suggesting that supervised signals derived from pretrained vision models can provide more semantically aligned guidance than textual conditions. 

\section{Discussions and Conclusion}

\noindent\textbf{Connection with dispersive loss.}
Dispersive loss~\cite{wang2025diffusedisperseimagegeneration} can be viewed as a special case of CARE, where $\psi(\cdot)$ is a constant mapping and thereby the conditional supervision is removed, obtaining an unsupervised contrastive loss that is condition-agnostic.

\noindent\textbf{Limitations and future work.}
Our study focuses in large part on categorical and textual conditions and is limited to diffusion models for image generation. Exploring the applicability of CARE to richer supervision, such as multimodal conditioning, or extending CARE to video, audio, or 3D content generative frameworks are promising directions for future work.

\noindent\textbf{Conclusion.}
We introduced CARE, a lightweight condition-aware regularization method that uses built-in conditioning signals to improve the semantic structure of diffusion representations. CARE is theoretically connected to mutual-information maximization and empirically enhances sample quality for both class-to-image and text-to-image tasks. It requires no external models and complements existing approaches, offering a distinct and synergistic source of improvement.

\section*{Impact Statement}
This paper presents work whose goal is to advance the field of Machine
Learning. There are many potential societal consequences of our work, none
which we feel must be specifically highlighted here.

% Authors are \textbf{required} to include a statement of the potential broader
% impact of their work, including its ethical aspects and future societal
% consequences. This statement should be in an unnumbered section at the end of
% the paper (co-located with Acknowledgements -- the two may appear in either
% order, but both must be before References), and does not count toward the paper
% page limit. In many cases, where the ethical impacts and expected societal
% implications are those that are well established when advancing the field of
% Machine Learning, substantial discussion is not required, and a simple
% statement such as the following will suffice:

% ``This paper presents work whose goal is to advance the field of Machine
% Learning. There are many potential societal consequences of our work, none
% which we feel must be specifically highlighted here.''

% The above statement can be used verbatim in such cases, but we encourage
% authors to think about whether there is content which does warrant further
% discussion, as this statement will be apparent if the paper is later flagged
% for ethics review.


\bibliography{example_paper}

@String(NIPS= {Adv. Neural Inform. Process. Syst.})

@String(NIPS  = {NeurIPS})

@inproceedings{huh2024position,
  title={Position: The platonic representation hypothesis},
  author={Huh, Minyoung and Cheung, Brian and Wang, Tongzhou and Isola, Phillip},
  booktitle={Forty-first International Conference on Machine Learning},
  year={2024}
}

@inproceedings{yu2024repa,
    title={Representation Alignment for Generation: Training Diffusion Transformers Is Easier Than You Think},
    author={Sihyun Yu and Sangkyung Kwak and Huiwon Jang and Jongheon Jeong and Jonathan Huang and Jinwoo Shin and Saining Xie},
    year={2025},
    booktitle={International Conference on Learning Representations},
}

@article{wang2025diffusedisperseimagegeneration,
  title={Diffuse and Disperse: Image Generation with Representation Regularization}, 
  author={Runqian Wang and Kaiming He},
  year={2025},
  eprint={2506.09027},
  journal={arXiv preprint arXiv:2506.09027},
  archivePrefix={arXiv},
  primaryClass={cs.CV},
  url={https://arxiv.org/abs/2506.09027}
}

@InProceedings{understanding-hypersphere,
  title = 	 {Understanding Contrastive Representation Learning through Alignment and Uniformity on the Hypersphere},
  author =       {Wang, Tongzhou and Isola, Phillip},
  booktitle = 	 {Proceedings of the 37th International Conference on Machine Learning},
  pages = 	 {9929--9939},
  year = 	 {2020},
  editor = 	 {III, Hal Daumé and Singh, Aarti},
  volume = 	 {119},
  series = 	 {Proceedings of Machine Learning Research},
  month = 	 {13--18 Jul},
  publisher =    {PMLR},
  url = 	 {https://proceedings.mlr.press/v119/wang20k.html}
}

@misc{wu2025representationentanglementgenerationtraining,
      title={Representation Entanglement for Generation: Training Diffusion Transformers Is Much Easier Than You Think}, 
      author={Ge Wu and Shen Zhang and Ruijing Shi and Shanghua Gao and Zhenyuan Chen and Lei Wang and Zhaowei Chen and Hongcheng Gao and Yao Tang and Jian Yang and Ming-Ming Cheng and Xiang Li},
      year={2025},
      eprint={2507.01467},
      archivePrefix={arXiv},
      primaryClass={cs.CV},
      url={https://arxiv.org/abs/2507.01467}, 
}

@misc{chen2025sarastructuraladversarialrepresentation,
      title={SARA: Structural and Adversarial Representation Alignment for Training-efficient Diffusion Models}, 
      author={Hesen Chen and Junyan Wang and Zhiyu Tan and Hao Li},
      year={2025},
      eprint={2503.08253},
      archivePrefix={arXiv},
      primaryClass={cs.CV},
      url={https://arxiv.org/abs/2503.08253}, 
}

@article{fid,
  title={Gans trained by a two time-scale update rule converge to a local nash equilibrium},
  author={Heusel, Martin and Ramsauer, Hubert and Unterthiner, Thomas and Nessler, Bernhard and Hochreiter, Sepp},
  journal={Advances in neural information processing systems},
  volume={30},
  year={2017}
}

@inproceedings{sohl,
  title={Deep unsupervised learning using nonequilibrium thermodynamics},
  author={Sohl-Dickstein, Jascha and Weiss, Eric and Maheswaranathan, Niru and Ganguli, Surya},
  booktitle={International conference on machine learning},
  pages={2256--2265},
  year={2015},
  organization={pmlr}
}

@article{ddpm,
  title={Denoising diffusion probabilistic models},
  author={Ho, Jonathan and Jain, Ajay and Abbeel, Pieter},
  journal={Advances in neural information processing systems},
  volume={33},
  pages={6840--6851},
  year={2020}
}

@article{song2020score,
  title={Score-based generative modeling through stochastic differential equations},
  author={Song, Yang and Sohl-Dickstein, Jascha and Kingma, Diederik P and Kumar, Abhishek and Ermon, Stefano and Poole, Ben},
  journal={arXiv preprint arXiv:2011.13456},
  year={2020}
}

@inproceedings{
song2021denoising,
title={Denoising Diffusion Implicit Models},
author={Jiaming Song and Chenlin Meng and Stefano Ermon},
booktitle={International Conference on Learning Representations},
year={2021},
url={https://openreview.net/forum?id=St1giarCHLP}
}

@article{yang2024cogvideox,
  title={Cogvideox: Text-to-video diffusion models with an expert transformer},
  author={Yang, Zhuoyi and Teng, Jiayan and Zheng, Wendi and Ding, Ming and Huang, Shiyu and Xu, Jiazheng and Yang, Yuanming and Hong, Wenyi and Zhang, Xiaohan and Feng, Guanyu and others},
  journal={arXiv preprint arXiv:2408.06072},
  year={2024}
}

@article{wan2025wan,
  title={Wan: Open and advanced large-scale video generative models},
  author={Wan, Team and Wang, Ang and Ai, Baole and Wen, Bin and Mao, Chaojie and Xie, Chen-Wei and Chen, Di and Yu, Feiwu and Zhao, Haiming and Yang, Jianxiao and others},
  journal={arXiv preprint arXiv:2503.20314},
  year={2025}
}

@inproceedings{esser2024scaling,
  title={Scaling rectified flow transformers for high-resolution image synthesis},
  author={Esser, Patrick and Kulal, Sumith and Blattmann, Andreas and Entezari, Rahim and M{\"u}ller, Jonas and Saini, Harry and Levi, Yam and Lorenz, Dominik and Sauer, Axel and Boesel, Frederic and others},
  booktitle={Forty-first international conference on machine learning},
  year={2024}
}

@article{betker2023improving,
  title={Improving image generation with better captions},
  author={Betker, James and Goh, Gabriel and Jing, Li and Brooks, Tim and Wang, Jianfeng and Li, Linjie and Ouyang, Long and Zhuang, Juntang and Lee, Joyce and Guo, Yufei and others},
  journal={Computer Science. https://cdn. openai. com/papers/dall-e-3. pdf},
  volume={2},
  number={3},
  pages={8},
  year={2023}
}

@misc{flux2024,
    author={Black Forest Labs},
    title={FLUX},
    year={2024},
    howpublished={\url{https://github.com/black-forest-labs/flux}},
}

@misc{hunyuan3d22025tencent,
    title={Hunyuan3D 2.0: Scaling Diffusion Models for High Resolution Textured 3D Assets Generation},
    author={Tencent Hunyuan3D Team},
    year={2025},
    eprint={2501.12202},
    archivePrefix={arXiv},
    primaryClass={cs.CV}
}

@article{dhariwal2021diffusion,
  title={Diffusion models beat gans on image synthesis},
  author={Dhariwal, Prafulla and Nichol, Alexander},
  journal={Advances in neural information processing systems},
  volume={34},
  pages={8780--8794},
  year={2021}
}

@inproceedings{
    lipman2023flow,
    title={Flow Matching for Generative Modeling},
    author={Yaron Lipman and Ricky T. Q. Chen and Heli Ben-Hamu and Maximilian Nickel and Matthew Le},
    booktitle={The Eleventh International Conference on Learning Representations },
    year={2023},
    url={https://openreview.net/forum?id=PqvMRDCJT9t}
}

@inproceedings{sit,
  title={{SiT}: Exploring flow and diffusion-based generative models with scalable interpolant transformers},
  author={Ma, Nanye and Goldstein, Mark and Albergo, Michael S and Boffi, Nicholas M and Vanden-Eijnden, Eric and Xie, Saining},
  booktitle={European Conference on Computer Vision},
  pages={23--40},
  year={2024},
  organization={Springer}
}

@inproceedings{imgnet,
  title={{ImageNet}: A large-scale hierarchical image database},
  author={Deng, Jia and Dong, Wei and Socher, Richard and Li, Li-Jia and Li, Kai and Fei-Fei, Li},
  booktitle={2009 IEEE conference on computer vision and pattern recognition},
  pages={248--255},
  year={2009},
  organization={Ieee}
}

@inproceedings{ldm,
  title={High-resolution image synthesis with latent diffusion models},
  author={Rombach, Robin and Blattmann, Andreas and Lorenz, Dominik and Esser, Patrick and Ommer, Bj{\"o}rn},
  booktitle={Proceedings of the IEEE/CVF conference on computer vision and pattern recognition},
  pages={10684--10695},
  year={2022}
}

@article{cfg,
  title={Classifier-free diffusion guidance},
  author={Ho, Jonathan and Salimans, Tim},
  journal={arXiv preprint arXiv:2207.12598},
  year={2022}
}

@inproceedings{llava,
 author = {Liu, Haotian and Li, Chunyuan and Wu, Qingyang and Lee, Yong Jae},
 booktitle = {Advances in Neural Information Processing Systems},
 editor = {A. Oh and T. Naumann and A. Globerson and K. Saenko and M. Hardt and S. Levine},
 pages = {34892--34916},
 publisher = {Curran Associates, Inc.},
 title = {Visual Instruction Tuning},
 url = {https://proceedings.neurips.cc/paper_files/paper/2023/file/6dcf277ea32ce3288914faf369fe6de0-Paper-Conference.pdf},
 volume = {36},
 year = {2023}
}

@inproceedings{cc3m,
    title = "Conceptual Captions: A Cleaned, Hypernymed, Image Alt-text Dataset For Automatic Image Captioning",
    author = "Sharma, Piyush  and
      Ding, Nan  and
      Goodman, Sebastian  and
      Soricut, Radu",
    editor = "Gurevych, Iryna  and
      Miyao, Yusuke",
    booktitle = "Proceedings of the 56th Annual Meeting of the Association for Computational Linguistics (Volume 1: Long Papers)",
    month = jul,
    year = "2018",
    address = "Melbourne, Australia",
    publisher = "Association for Computational Linguistics",
    url = "https://aclanthology.org/P18-1238/",
    doi = "10.18653/v1/P18-1238",
    pages = "2556--2565"
}

@inproceedings{clip,
  author       = {Alec Radford and
                  Jong Wook Kim and
                  Chris Hallacy and
                  Aditya Ramesh and
                  Gabriel Goh and
                  Sandhini Agarwal and
                  Girish Sastry and
                  Amanda Askell and
                  Pamela Mishkin and
                  Jack Clark and
                  Gretchen Krueger and
                  Ilya Sutskever},
  editor       = {Marina Meila and
                  Tong Zhang},
  title        = {Learning Transferable Visual Models From Natural Language Supervision},
  booktitle    = {Proceedings of the 38th International Conference on Machine Learning,
                  {ICML} 2021, 18-24 July 2021, Virtual Event},
  series       = {Proceedings of Machine Learning Research},
  volume       = {139},
  pages        = {8748--8763},
  publisher    = {{PMLR}},
  year         = {2021},
  url          = {http://proceedings.mlr.press/v139/radford21a.html},
  bibsource    = {dblp computer science bibliography, https://dblp.org}
}

@inproceedings{clipscore,
    title = "{CLIPS}core: A Reference-free Evaluation Metric for Image Captioning",
    author = "Hessel, Jack  and
      Holtzman, Ari  and
      Forbes, Maxwell  and
      Le Bras, Ronan  and
      Choi, Yejin",
    editor = "Moens, Marie-Francine  and
      Huang, Xuanjing  and
      Specia, Lucia  and
      Yih, Scott Wen-tau",
    booktitle = "Proceedings of the 2021 Conference on Empirical Methods in Natural Language Processing",
    month = nov,
    year = "2021",
    address = "Online and Punta Cana, Dominican Republic",
    publisher = "Association for Computational Linguistics",
    url = "https://aclanthology.org/2021.emnlp-main.595/",
    doi = "10.18653/v1/2021.emnlp-main.595",
    pages = "7514--7528"
}

@misc{oquab2023dinov2,
  title={DINOv2: Learning Robust Visual Features without Supervision},
  author={Oquab, Maxime and Darcet, Timothée and Moutakanni, Theo and Vo, Huy V. and Szafraniec, Marc and Khalidov, Vasil and Fernandez, Pierre and Haziza, Daniel and Massa, Francisco and El-Nouby, Alaaeldin and Howes, Russell and Huang, Po-Yao and Xu, Hu and Sharma, Vasu and Li, Shang-Wen and Galuba, Wojciech and Rabbat, Mike and Assran, Mido and Ballas, Nicolas and Synnaeve, Gabriel and Misra, Ishan and Jegou, Herve and Mairal, Julien and Labatut, Patrick and Joulin, Armand and Bojanowski, Piotr},
  journal={arXiv:2304.07193},
  year={2023}
}

@misc{chen2025carflowconditionawarereparameterizationaligns,
      title={CAR-Flow: Condition-Aware Reparameterization Aligns Source and Target for Better Flow Matching}, 
      author={Chen Chen and Pengsheng Guo and Liangchen Song and Jiasen Lu and Rui Qian and Xinze Wang and Tsu-Jui Fu and Wei Liu and Yinfei Yang and Alex Schwing},
      year={2025},
      eprint={2509.19300},
      archivePrefix={arXiv},
      primaryClass={cs.CV},
      url={https://arxiv.org/abs/2509.19300}, 
}

@misc{issachar2025designingconditionalpriordistribution,
      title={Designing a Conditional Prior Distribution for Flow-Based Generative Models}, 
      author={Noam Issachar and Mohammad Salama and Raanan Fattal and Sagie Benaim},
      year={2025},
      eprint={2502.09611},
      archivePrefix={arXiv},
      primaryClass={cs.LG},
      url={https://arxiv.org/abs/2502.09611}, 
}

@inproceedings{
    liu2023rectified-flow,
    title={Flow Straight and Fast: Learning to Generate and Transfer Data with Rectified Flow},
    author={Xingchao Liu and Chengyue Gong and qiang liu},
    booktitle={The Eleventh International Conference on Learning Representations },
    year={2023},
    url={https://openreview.net/forum?id=XVjTT1nw5z}
}

@InProceedings{Huynh_2022_WACV_false_negative_cancellation,
    author    = {Huynh, Tri and Kornblith, Simon and Walter, Matthew R. and Maire, Michael and Khademi, Maryam},
    title     = {Boosting Contrastive Self-Supervised Learning With False Negative Cancellation},
    booktitle = {Proceedings of the IEEE/CVF Winter Conference on Applications of Computer Vision (WACV)},
    month     = {January},
    year      = {2022},
    pages     = {2785-2795}
}

@article{nadaraya-kernel-regression,
author = {Nadaraya, E. A.},
title = {On Estimating Regression},
journal = {Theory of Probability \& Its Applications},
volume = {9},
number = {1},
pages = {141-142},
year = {1964},
doi = {10.1137/1109020},

URL = {https://doi.org/10.1137/1109020},
eprint = {https://doi.org/10.1137/1109020}
}

@article{watson-kernel-regression,
 ISSN = {0581572X},
 URL = {http://www.jstor.org/stable/25049340},
 author = {Geoffrey S. Watson},
 journal = {Sankhyā: The Indian Journal of Statistics, Series A (1961-2002)},
 number = {4},
 pages = {359--372},
 publisher = {Springer},
 title = {Smooth Regression Analysis},
 urldate = {2025-11-13},
 volume = {26},
 year = {1964}
}

@inproceedings{salimans-inception-score,
author = {Salimans, Tim and Goodfellow, Ian and Zaremba, Wojciech and Cheung, Vicki and Radford, Alec and Chen, Xi},
title = {Improved techniques for training GANs},
year = {2016},
isbn = {9781510838819},
publisher = {Curran Associates Inc.},
address = {Red Hook, NY, USA},
booktitle = {Proceedings of the 30th International Conference on Neural Information Processing Systems},
pages = {2234–2242},
numpages = {9},
location = {Barcelona, Spain},
series = {NIPS'16}
}
\bibliographystyle{icml2026}

%%%%%%%%%%%%%%%%%%%%%%%%%%%%%%%%%%%%%%%%%%%%%%%%%%%%%%%%%%%%%%%%%%%%%%%%%%%%%%%
%%%%%%%%%%%%%%%%%%%%%%%%%%%%%%%%%%%%%%%%%%%%%%%%%%%%%%%%%%%%%%%%%%%%%%%%%%%%%%%
% APPENDIX
%%%%%%%%%%%%%%%%%%%%%%%%%%%%%%%%%%%%%%%%%%%%%%%%%%%%%%%%%%%%%%%%%%%%%%%%%%%%%%%
%%%%%%%%%%%%%%%%%%%%%%%%%%%%%%%%%%%%%%%%%%%%%%%%%%%%%%%%%%%%%%%%%%%%%%%%%%%%%%%
\newpage
\appendix
\onecolumn

\section{Heuristic Interpretation of CARE via Mutual Information}
\label{app:interpretation}

In this appendix, we provide a heuristic interpretation of the CARE objective
from the perspective of conditional mutual information under the class-conditional setting. 
We emphasize that this analysis is \emph{not} required for defining or optimizing CARE;
rather, it serves to offer intuition on how the proposed regularizer encourages
structured conditional representations.

\paragraph{From mutual information to conditional likelihood}

Given a batch of representation--condition pairs
$\{(\mathbf{z}_i, \mathbf{c}_i)\}_{i=1}^N$. 
From a probabilistic perspective, this behavior can be interpreted as encouraging
dependence between the representation variable $Z_\theta$ and the condition variable $C$.
A natural quantity that captures such dependence is the mutual information
\begin{equation}
    I(Z_\theta; C) = H(C) - H(C \mid Z_\theta),
\end{equation}
where $Z_\theta$ and $C$ follow the joint distribution $p_\theta(z,c)$ induced by the data
and the model.
Since $H(C)$ is independent of model parameters, maximizing $I(Z_\theta; C)$
is equivalent to minimizing the conditional entropy $H(C \mid Z_\theta)$,
or equivalently maximizing
\begin{equation}
    \mathbb{E}_{p_\theta(z,c)} \big[ \log p(C \mid Z_\theta) \big].
\end{equation}

\paragraph{Kernel-based surrogate for the conditional likelihood.}
Directly optimizing the conditional likelihood $p(C \mid Z_\theta)$ is intractable,
as this distribution is implicit and evolves with model parameters.
Instead, we seek a tractable surrogate objective that captures the same inductive bias:
representations corresponding to similar conditions should exhibit stronger geometric coherence.

To this end, we adopt a nonparametric kernel-based approximation of
$p(C \mid Z_\theta)$ using Nadaraya--Watson regression~\cite{nadaraya-kernel-regression, watson-kernel-regression}.
For clarity, we focus on discrete conditions.
Letting $X = Z_\theta$ and $Y = \mathds{1}\{C = \mathbf{c}\}$,
the conditional probability can be estimated from a batch as
\begin{equation}
    \hat{p}(\mathbf{c} \mid \mathbf{z})
    = \frac{\sum_{i=1}^{N} \phi(\mathbf{z}, \mathbf{z}_i)\,
    \mathds{1}\{\mathbf{c}_i = \mathbf{c}\}}
    {\sum_{i=1}^{N} \phi(\mathbf{z}, \mathbf{z}_i)},
\end{equation}
where $\phi(\cdot,\cdot)$ is a kernel defined in the representation space.

To avoid numerical issues, we use a smoothed estimation 
\begin{align}
    \tilde{p}(\mathbf{c} \mid \mathbf{z}) = (1-\epsilon)\hat{p}(\mathbf{c} \mid \mathbf{z}) + \epsilon / K,
\end{align}
where $\epsilon > 0$ is a small constant and $K$ is the total number of classes.

\paragraph{A tractable surrogate objective.}
Using the kernel-based estimate of $p(C \mid Z_\theta)$,
one can construct a negative log-likelihood objective over a batch,
which takes the form
\begin{equation}
\mathcal{L}
= - \sum_{i=1}^{N} \log \tilde{p}(\mathbf{c}_i \mid \mathbf{z}_i),
\end{equation}

To obtain a compact and tractable form, we follow a common approximation
used in contrastive learning~\cite{understanding-hypersphere},
which swaps the order of the summation and the logarithm.
Although this approximation is not exact, it leads to a decomposed objective
consisting of a supervised, condition-aware term and an unsupervised
uniformity regularization term ($\mathcal{L}_{\text{uniformity}}$, as in \citet{understanding-hypersphere}).
% Importantly, CARE does not rely on the exactness of this approximation;
% rather, it inherits the same inductive bias widely used in representation learning.

\begin{align}
    \mathcal{L}_{\text{nll}} = & { - \log \left( { \sum_{1\leq i,j \leq N} W_{i,j} \phi\left(\mathbf{z}_i, \mathbf{z}_j\right) } \right) } \nonumber
     + \underbrace{ \log \left( \sum_{1 \leq i,j \leq N} \phi\left(\mathbf{z}_i, \mathbf{z}_j\right) \right) }_{ \mathcal{L}_\text{uniformity}}, 
\end{align}
where $W_{i,j} = (1-\epsilon)\mathds{1}\left\{c_i = c_j\right\} + \epsilon / K$. 

To explicitly control the balance between these two effects,
we introduce a coefficient $\gamma$ and obtain the surrogate objective
\begin{equation}
\mathcal{L}_{\gamma}
= \mathcal{L}_{\text{nll}} + \gamma\, \mathcal{L}_{\text{uniformity}},
\label{eq:l-gamma}
\end{equation}

% We stress that $\mathcal{L}_{\gamma}$ should be viewed as a convenient
% regularized surrogate, rather than an exact estimator of mutual information.

\paragraph{Simplified form under class-conditional structure.}
For discrete conditions, let
\[
X = \sum_{\mathbf{c}_i = \mathbf{c}_j} \phi(\mathbf{z}_i, \mathbf{z}_j),
\qquad
Y = \sum_{\mathbf{c}_i \neq \mathbf{c}_j} \phi(\mathbf{z}_i, \mathbf{z}_j).
\]
With a smoothing constant $\beta$ induced by label smoothing,
the objective $\mathcal{L}_{\gamma}$ can be rewritten (up to an additive constant) as
\begin{equation}
\mathcal{L}_{\beta,\gamma}
= - \log\left( X + \beta Y \right)
+ (1 + \gamma)\log\left( X + Y \right).
\end{equation}

% \paragraph{Relation to InfoNCE-style objectives.}

\begin{figure}[h]
    \centering
    \includegraphics[width=0.4\linewidth]{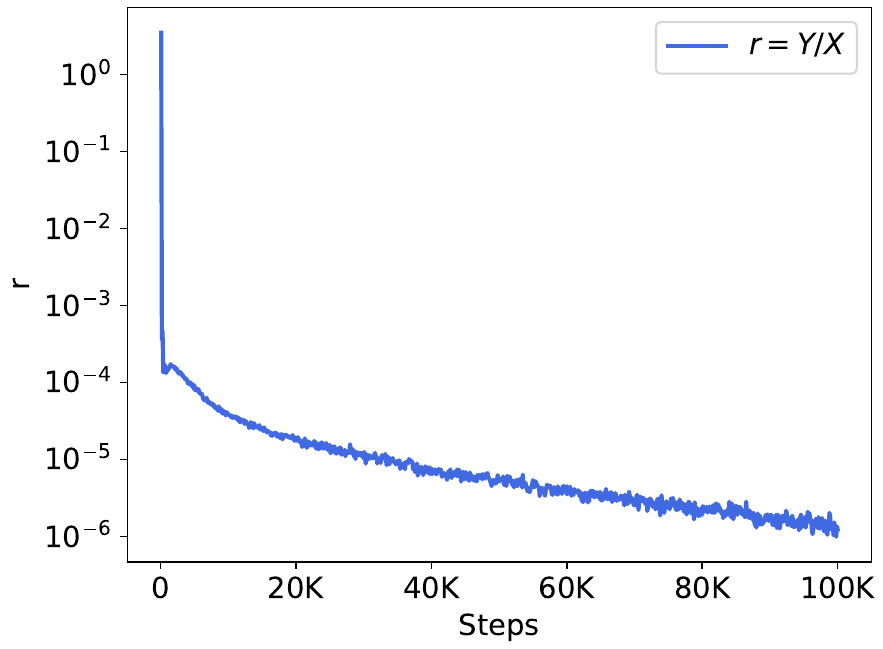}
    \caption{Evolution of the ratio $r = Y / X$ over training steps of CARE. The ratio quickly decreases to the order of $10^{-4}$ after several hundreds steps.}
    \label{fig:plot-r}
\end{figure}

\paragraph{Resulting CARE form.}
Under the empirical observation that the ratio $r = Y / X$
quickly becomes small during training (see \cref{fig:plot-r}),
$\mathcal{L}_{\beta,\gamma}$ admits a first-order approximation
that leads to a concise form
\begin{equation}\label{eq:l-alpha}
\mathcal{L}
= \log \left( \alpha X + Y \right), \qquad 0 < \alpha < 1,
\end{equation}
where $\alpha$ absorbs the effects of $\gamma$ and $\beta$ as follows
\begin{align}
    \mathcal{L}_{\beta, \gamma} 
    & = 
    \gamma \log X + \left( 1 + \gamma - \beta \right) r + \mathcal{O}(r^2)\\
    & = \gamma \log \left( \frac{\gamma}{1 + \gamma - \beta} X + Y\right) + C + \mathcal{O}(r^2). \label{constant-in-loss}
\end{align} 

\cref{eq:l-alpha} is exactly the formalization of CARE under class-conditioned settings (see ~\cref{sec:instantiation-class-cond}). 

\newpage
\section{Additional Visual Results on ImageNet 256 $\times$ 256}
\begin{figure}[h]
    \centering
    \includegraphics[width=0.8\linewidth]{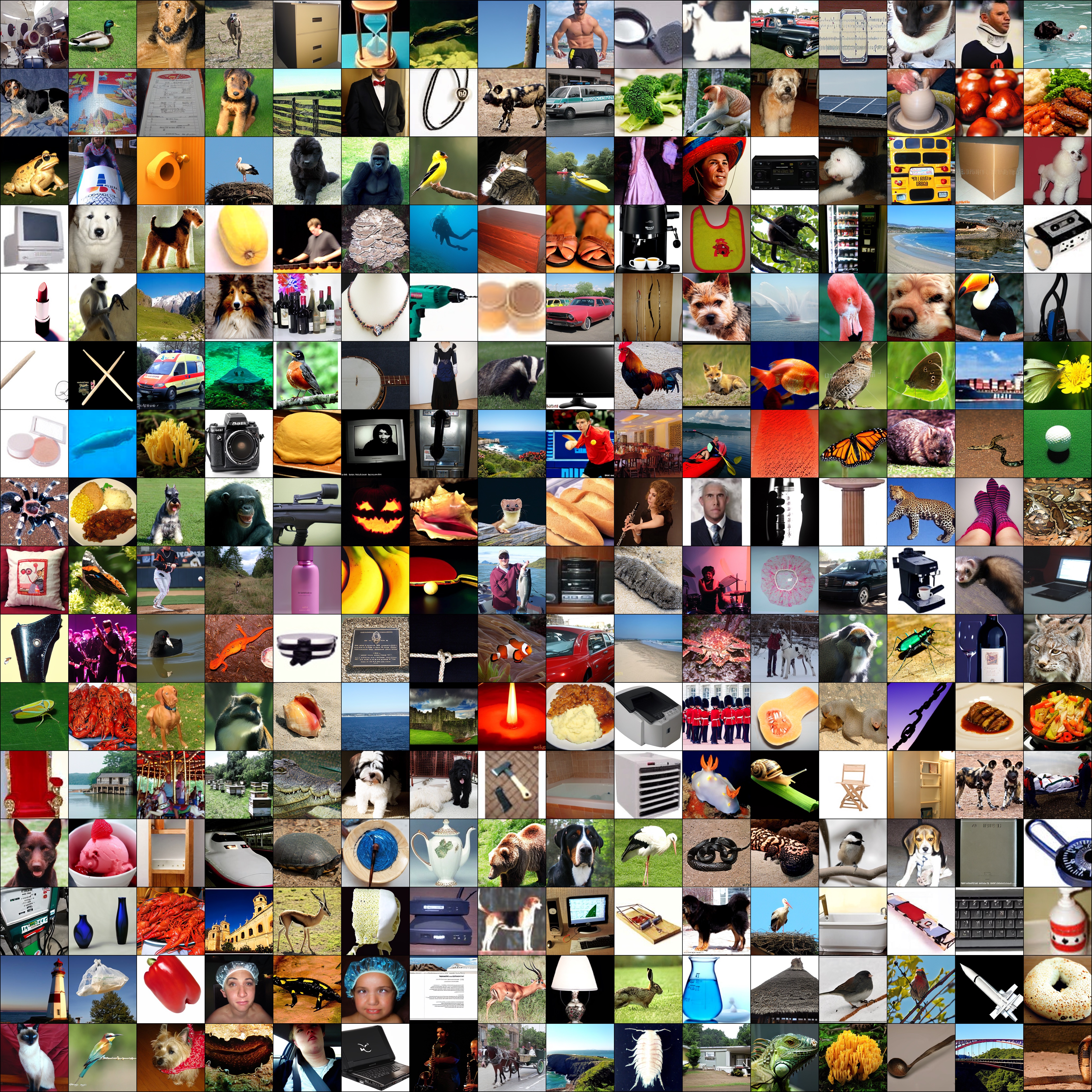}
    \caption{Uncurated samples generated by SiT-XL/2 trained with CARE for 2.4M iterations. Best view zoom in. Sampling uses a 50-step ODE sampler with CFG scale 4.0. }
    \label{fig:more-samples}
\end{figure}


% You can have as much text here as you want. The main body must be at most $8$
% pages long. For the final version, one more page can be added. If you want, you
% can use an appendix like this one.

% The $\mathtt{\backslash onecolumn}$ command above can be kept in place if you
% prefer a one-column appendix, or can be removed if you prefer a two-column
% appendix.  Apart from this possible change, the style (font size, spacing,
% margins, page numbering, etc.) should be kept the same as the main body.
%%%%%%%%%%%%%%%%%%%%%%%%%%%%%%%%%%%%%%%%%%%%%%%%%%%%%%%%%%%%%%%%%%%%%%%%%%%%%%%
%%%%%%%%%%%%%%%%%%%%%%%%%%%%%%%%%%%%%%%%%%%%%%%%%%%%%%%%%%%%%%%%%%%%%%%%%%%%%%%

\end{document}